\documentclass[letterpaper, 10pt, conference]{ieeeconf}
\IEEEoverridecommandlockouts
\usepackage{cite}
\usepackage{amsmath,amssymb,amsfonts}
\usepackage{graphicx}
\usepackage{xcolor}

\usepackage{algorithm}
\usepackage{algpseudocode}
\usepackage{comment}
  \usepackage{amsthm}

\newtheorem{proposition}{Proposition}
\theoremstyle{remark}

\theoremstyle{definition}

\newtheorem{theorem}{Theorem}

\title{\LARGE \bf Mobility-Budget-Constrained Submodular Maximization for Distributed Semantic Multi-Robot Exploration}

\author{Changxuan Yao$^{1,*}$, Sandilya Sai Garimella$^{1,*}$, Lu Gan$^{1}$, and Panagiotis Tsiotras$^{1}$%
\thanks{$^{*}$These authors contributed equally.}%
\thanks{$^{1}$The authors are with the Institute of Robotics and Intelligent Machines, Georgia Institute of Technology, Atlanta, GA 30332, USA.
{\tt\small \{cyao48, sgarimella34, lgan, tsiotras\}@gatech.edu}}%
}

\begin{document}

\maketitle
\thispagestyle{empty}
\pagestyle{empty}

\begin{abstract}
Multi-robot exploration under communication and mobility constraints requires balancing information gain with reachability. We present a distributed semantic exploration framework that combines Dirichlet mapping, budget-constrained trajectory selection, and Resource-Aware distributed Greedy (RAG) coordination. Each robot clusters frontiers, computes terrain- and heading-aware least-cost paths with a single budget-limited Dijkstra search, and selects feasible paths by marginal information gain, represented by the expected reduction in Dirichlet entropy. Each robot's per-round selection is exact, the team's joint per-round information gain is at least half the centralized optimum, and the bound degrades gracefully under limited communication. Ten-run gridworld ablations show improved coverage, semantic mIoU, and map certainty with lower entropy and traversal cost per unit distance than shortest-distance frontier selection. Budget-sensitivity experiments show that an intermediate budget best balances path reach and receding-horizon execution. In photorealistic full-system simulation against five coordination rules run in their native published forms, the proposed method attains the highest mean coverage and semantic accuracy while being the only one that respects the per-trajectory mobility budget.
\end{abstract}

\section{Introduction}

\begin{figure}[t]
\centering
\includegraphics[width=\linewidth]{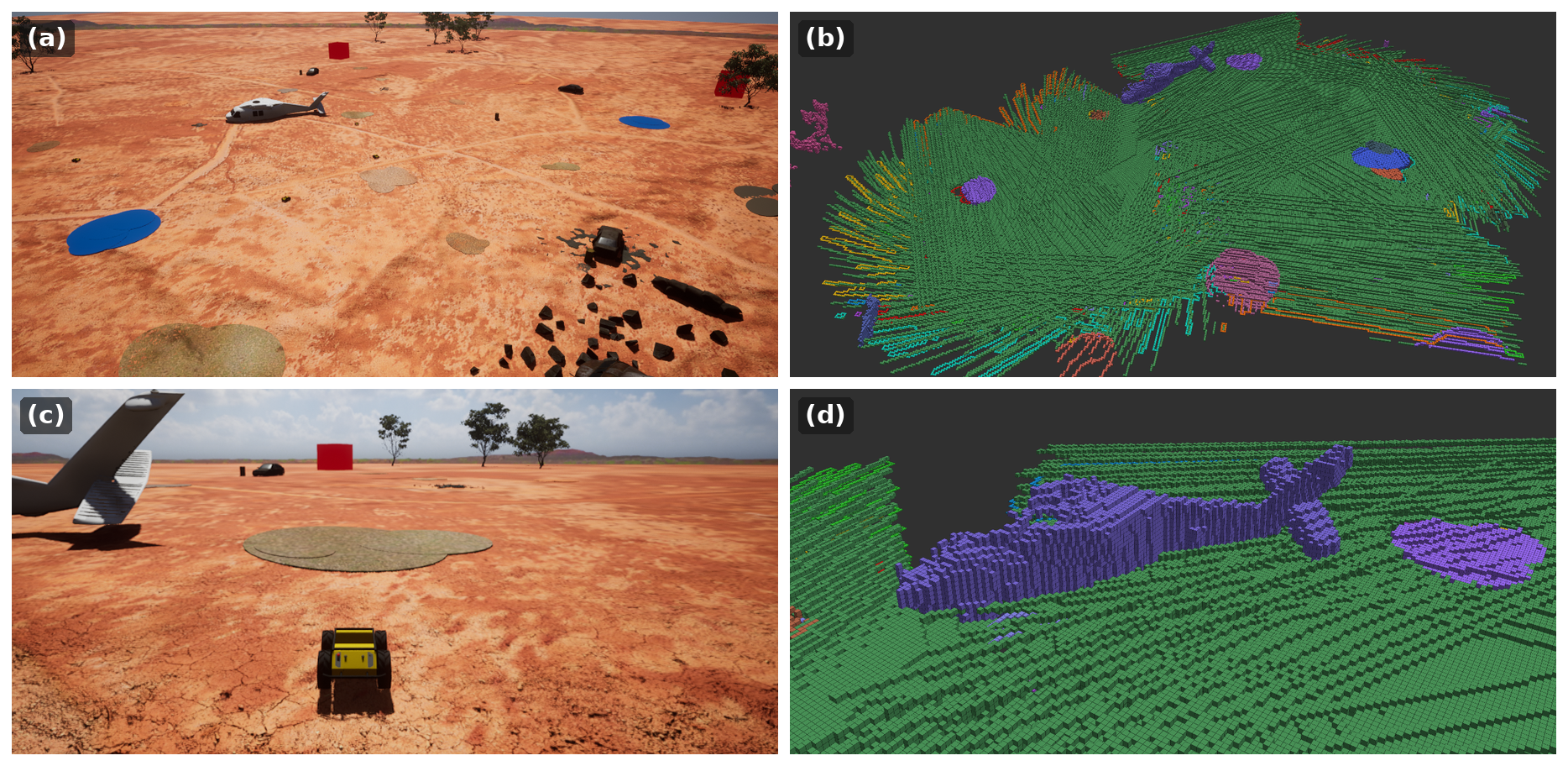}
\caption{Distributed semantic multi-robot exploration in HERCULES. (a) The outdoor arena, with terrain patches of differing traversability alongside structures and debris. (b) The 3-D semantic voxel map accumulated online by the team, colored by inferred class. (c) A Husky ground robot exploring the arena. (d) A reconstructed aircraft and adjacent terrain patch.}
\label{fig:hero}
\end{figure}

Multi-robot exploration enables teams to map unknown environments for applications such as search and rescue, reconnaissance, and planetary exploration \cite{collab_slam_survey}. Robots must coordinate their actions while balancing information acquisition against resource limitations. Semantic maps require more communication bandwidth than binary occupancy maps, which can delay and degrade coordination, so resource-aware semantic information acquisition is essential (Fig.~\ref{fig:hero}).

We propose a distributed semantic exploration framework in which each robot maintains a Dirichlet semantic map fused from source-attributed records exchanged with neighbors. Frontier regions are clustered into candidate targets, and a single Dijkstra search over a heading-augmented graph returns least-cost paths under semantic traversal and heading-change costs, limited by a per-trajectory mobility budget that reflects the energy and time limits of SWaP-constrained ground robots. Robots de-conflict trajectories through the Resource-Aware distributed Greedy (RAG) protocol \cite{xu2025communication} using local marginal information gain, represented by the expected reduction in Dirichlet entropy.

The contributions of this work are:
\begin{enumerate}
\item A mobility-budget-constrained submodular formulation that maximizes semantic information gain subject to an exact, edge-additive mobility budget incorporating terrain-dependent traversal and heading-change costs.
\item RAG-based distributed coordination with approximation guarantees under limited communication.
\item A complete multi-robot implementation in MATLAB and high-fidelity ground-robot simulations, evaluated through ten-run paired gridworld ablations, mobility-budget sensitivity experiments, and comparisons against five coordination rules in both matched framework and native published forms.
\end{enumerate}

\section{Related Work}

Communication-constrained multi-robot exploration includes distributed SLAM \cite{kimera-multi}, low-bandwidth planning \cite{comm-effi-large-env}, and dynamic relay architectures \cite{dynamic_hier}; extreme-environment deployments further motivate resource-efficient designs \cite{darpa-subt, chord, lamp}. Active exploration methods select trajectories using information-theoretic objectives \cite{active-slam-survey, dec-info-acq, dec-int-het, active-slam-ent-op}. Bayesian multi-class mapping maintains per-cell categorical distributions and provides a closed-form semantic mutual-information bound \cite{bayesian-mapping}, supporting single-robot semantic exploration \cite{ssmi} and multi-robot distributed optimization \cite{asgharivaskasi2025riemannian}. These methods do not jointly address terrain-dependent mobility budgets and limited inter-robot communication. RAG \cite{xu2025communication} enables communication-efficient distributed submodular maximization with topology-dependent guarantees. We combine a Dirichlet-categorical semantic grid with terrain-aware mobility constraints and RAG-based coordination under limited communication.

\section{Proposed Method}
\label{sec:method}

\subsection{Semantic Map Representation and Fusion}

Following \cite{bayesian-mapping}, the map $M$ is a 2-D grid where each cell $j$ carries a semantic probability vector $\theta_j = (\theta^1_j, \dots, \theta^K_j)$ over $K$ classes. Given one-hot semantic measurements and a Dirichlet prior $\text{Dir}(\boldsymbol{\alpha}_0)$, the posterior parameters of cell $j$ are
\begin{equation}
\label{eq:dir_posterior}
\alpha^k_{j} = \alpha^k_{0} + \sum_{x_i} k(x_i, x_j)\, y^k_{i},
\end{equation}
where the Gaussian kernel weight $k(x_i, x_j)$ encodes that closer observations are more reliable. The map provides the posterior mean $E[\theta^k_j] = \alpha^k_j / \sum_k \alpha^k_j$ and the differential entropy $H(\text{Dir}(\boldsymbol{\alpha}_j))$ \cite{lin2016dirichlet}. The mutual information between the map and future observations $Z$ over a perceptual field $\mathcal{O}$ is
\begin{equation}
\label{eq:mutual_info}
I(M;Z) = \sum_{j \in \mathcal{O}} \left[
H\bigl(\text{Dir}(\boldsymbol{\alpha}^{\text{prior}}_j)\bigr)
-
\mathbb{E}_{Z}\!\left[
H\bigl(\text{Dir}(\boldsymbol{\alpha}^{\text{post}}_j)\bigr)
\right]
\right].
\end{equation}
During planning, future observations are unknown; each robot therefore evaluates this conditional expected reduction in entropy given its observations so far.

Each robot stores occupancy and semantic records by observation source. Let $\boldsymbol{\alpha}^{(i,s)}$ be source $s$'s semantic record stored by robot $i$, let $\mathcal{S}_i$ be the available sources, and let $C^{(i,s)} = \sum_j \sum_{k=1}^{K} (\alpha_j^{k,(i,s)} - \alpha_0^k)$ be the accumulated observation count of that record. When neighbors $i$ and $r$ communicate, they exchange their source-attributed records, and for each source $s$ robot $i$ retains the record with the larger count, replacing $\boldsymbol{\alpha}^{(i,s)}$ by $\boldsymbol{\alpha}^{(r,s)}$ whenever $C^{(r,s)} > C^{(i,s)}$. Records remain indexed by their original source, and robot $i$ forms its fused map as
\begin{equation}
\label{eq:fusion}
\alpha_j^{k,\mathrm{fused},(i)}
=
\alpha_0^k+
\sum_{s\in\mathcal{S}_i}
\left(\alpha_j^{k,(i,s)}-\alpha_0^k\right).
\end{equation}
Replacing, rather than summing, alternative versions of the same source record prevents double counting while allowing information to propagate through successive encounters. Under limited communication, each robot plans using the most complete records it has received.

\subsection{Budget-Constrained Problem Formulation}

A trajectory $a_i = (p^i_1, \dots, p^i_{L_i})$ is an ordered cell sequence, and its observation set $Z_i(a_i)$ is the union of sensor-footprint cells swept along it. Each cell $p$ has a semantic traversal cost $c_{s,p} = \sum_k E[\theta^k_p]\, S_k$, with $S \geq 0$ the per-class cost vector (e.g., $S_{\text{road}} < S_{\text{grass}} < S_{\text{mud}}$). The augmented trajectory cost adds a heading-change penalty over interior waypoints,
\begin{equation}
\label{eq:cost}
c_s(a_i) = \sum_{p \in a_i} c_{s,p} + \lambda \sum_{\ell=2}^{L_i-1} \bigl(\delta^i_\ell / \pi\bigr)^2,
\qquad \lambda \ll \min_k S_k,
\end{equation}
where $\delta^i_\ell$ is the turning angle at waypoint $\ell$. The multi-robot exploration problem is then
\begin{equation}
\label{eq:problem}
\max_{\mathbf{a} \in \mathcal{A}} \; I\Bigl(M; \bigcup_{i=1}^{n} Z_i(a_i)\Bigr)
\quad \text{s.t.} \quad c_s(a_i) \leq B \;\; \forall i,
\end{equation}
a monotone submodular maximization under per-robot budget constraints; since $c_s$ appears only in the constraint, the augmented cost leaves monotonicity and submodularity intact. Because costs are additive over the fixed least-cost path to each candidate rather than optimized over orderings, the budget remains a feasibility constraint on individual trajectories rather than an orienteering \cite{chekuri2005recursive} objective.

\subsection{Distributed Planning with RAG Coordination}

At round $t$, robot $i$ clusters frontier cells (unexplored cells adjacent to known traversable space) into candidate targets. One budget-limited Dijkstra search over a heading-augmented graph returns the least-cost path from the robot to every candidate. Graph nodes are cell-heading pairs, and edge costs realize \eqref{eq:cost}. Terminating the search at cost $B$ yields exactly the feasible candidate set $\tilde{\mathcal{A}}_i^{(t)} = \{a : c_s(a) \leq B\}$.

Because robot $i$ commits exactly one trajectory per round from the enumerated feasible set, its subproblem is solved exactly by $a_i^* = \arg\max_{a \in \tilde{\mathcal{A}}_i^{(t)}} \Delta(a \mid \psi_i^{(t)}, C)$, where $\Delta$ is the conditional expected marginal gain given observations so far and the trajectories $C$ committed by neighbors. The budget acts as a feasibility filter on $\tilde{\mathcal{A}}_i^{(t)}$ rather than as a knapsack over subsets, so each robot's selection is optimal rather than approximate. Coordination follows the RAG protocol \cite{xu2025communication}: undecided robots exchange their best marginal gains, the robot whose gain is largest among its undecided neighbors commits and broadcasts its trajectory, and the rest recompute marginals on the residual objective, so robots in disjoint neighborhoods commit concurrently. Algorithm~\ref{alg:exploration} summarizes one planning round.

\begin{algorithm}[t]
\footnotesize
\caption{Budget-Constrained Semantic Exploration with RAG Coordination (robot $i$, round $t$)}
\label{alg:exploration}
\begin{algorithmic}[1]
\Require fused map $M_i^{(t)}$, pose $x_i^{(t)}$, budget $B$, class costs $S$, neighborhood $N_i$
\Ensure committed trajectory $a_i^*$
\State Extract and cluster frontier cells of $M_i^{(t)}$ into candidate targets
\State Run one budget-limited Dijkstra search from $x_i^{(t)}$ over the heading-augmented graph with costs \eqref{eq:cost}
\State $\tilde{\mathcal{A}}_i^{(t)} \leftarrow$ least-cost paths with $c_s(a) \leq B$; \; $C \leftarrow \emptyset$
\Repeat
  \State Compute $\Delta(a \mid \psi_i^{(t)}, C)$ for all $a \in \tilde{\mathcal{A}}_i^{(t)}$
  \State $a_i^* \leftarrow \arg\max_{a \in \tilde{\mathcal{A}}_i^{(t)}} \Delta(a \mid \psi_i^{(t)}, C)$; \; $g_i \leftarrow \Delta(a_i^*)$
  \State Exchange $g_i$ with undecided robots in $N_i$
  \If{$g_i$ largest among undecided robots in $N_i$ (ties by index)}
    \State Commit $a_i^*$ and broadcast to $N_i$
  \Else
    \State Receive newly committed trajectories into $C$
  \EndIf
\Until{robot $i$ has committed}
\State Execute $a_i^*$; update via \eqref{eq:dir_posterior}; exchange, merge, and fuse records via \eqref{eq:fusion}
\end{algorithmic}
\end{algorithm}

\subsection{Approximation Guarantees}
Let $\mathrm{OPT}$ denote the centralized solution to \eqref{eq:problem} using all fused-map information available at round $t$. Conditional independence of observations given the map, as in the Dirichlet-categorical model, makes the objective monotone and submodular \cite{krause2008near, cover1999elements}. Full derivations are deferred to the journal version.

\begin{proposition}[Exact per-robot subproblem]
    \label{prop:exact}
    $\tilde{\mathcal{A}}_i^{(t)}$ is finite and explicitly enumerated by the budget-limited search, and robot $i$ commits a single trajectory per round. The selection above therefore attains the maximum of $\Delta$ over $\tilde{\mathcal{A}}_i^{(t)}$: the subproblem is solved exactly, $\alpha = 1$.
\end{proposition}

\begin{theorem}[Per-round, multi-robot]
\label{thm:multi}
With sequential commitment, full information exchange, and an $\alpha$-approximate marginal solver, the joint solution satisfies $f(\mathbf{a}) \geq \frac{\alpha}{1+\alpha} f(\mathrm{OPT})$.
\end{theorem}

Here $f$ is the per-round objective in \eqref{eq:problem}, the expected semantic information gain of the team's committed trajectories, and $\mathrm{OPT}$ the best joint assignment under the same budgets. With exact per-robot selection ($\alpha = 1$ by Proposition~\ref{prop:exact}), the team's joint per-round information gain is at least $\tfrac{1}{2}$ of $f(\mathrm{OPT})$. The per-robot problem is not a knapsack instance, since each robot commits one trajectory and infeasible trajectories are removed before selection. Under limited communication, RAG \cite{xu2025communication} gives
\begin{equation}
\label{eq:rag_bound}
f(\mathbf{a}^{\mathrm{RAG}})
\geq
\frac{1}{1+\kappa_f}
\left[
f(\mathrm{OPT})
-\kappa_f\sum_i \mathrm{coin}_{f,i}(N_i)
\right],
\end{equation}
where $\kappa_f$ is the curvature of $f$, and $\mathrm{coin}_{f,i}$ measures sensing overlap between robot $i$ and robots outside $N_i$. The degradation vanishes when non-neighbor footprints do not overlap, recovering the fully connected case when $\mathrm{coin}_{f,i}=0$. Map evidence still propagates transitively through source-attributed records, so limited communication delays map evidence but confines coordination to $N_i$.

\section{Experiments}

\subsection{Semantic Gridworld Ablation}

We evaluate the framework in a semantic gridworld containing six traversable terrain classes (dry soil, grass, dirt, sand, mud, and gravel) and obstacles. Five robots explore the initially unknown environment with sensor range $r_s=10$ cells, communication range $r_c=20$ cells, normalized mobility budget $B=1$, and a 5-s receding execution horizon. We use a symmetric Dirichlet prior $\alpha_0=1$, Gaussian observation kernel $\sigma=7$, terrain-cost vector $S=[0.09,0.12,0.14,0.18,0.22,0.33]$, and heading-change weight $\lambda=0.005$. The simulated sensor reports the correct class with probability $0.7$ and distributes the remaining probability uniformly among the incorrect classes.

We compare four modes: \textit{Shortest Distance}, nearest-frontier selection without IG or coordination; \textit{Geometric Cost + Semantic IG}; \textit{Semantic Cost + Occupancy IG}; and \textit{Full}, which combines semantic costs and semantic IG. The latter three use RAG coordination. We conduct 10 paired 300-s runs with identical seeds, initial positions, and headings across modes. Coverage is measured over traversable cells, while mIoU and Dirichlet differential entropy include unexplored traversable cells. Route cost is the standardized traversal and heading-change cost divided by total geometric distance.

\begin{table}[t]
\centering
\caption{Gridworld results over 10 paired runs (mean $\pm$ 95\% CI).}
\label{tab:ablation}
\setlength{\tabcolsep}{3pt}
\resizebox{\columnwidth}{!}{
\begin{tabular}{lcccc}
\hline
Mode & Cov. (\%) $\uparrow$ & mIoU (\%) $\uparrow$ & Entropy $\downarrow$ & Cost/dist. $\downarrow$\\
\hline
Shortest
& $79.9\pm7.0$ & $68.3\pm8.3$ & $-6.98\pm0.21$ & $0.113\pm0.004$\\
Geo.+Sem. IG
& $86.5\pm2.7$ & $73.6\pm4.0$ & $-7.37\pm0.08$ & $0.122\pm0.004$\\
Sem.+Occ. IG
& $85.3\pm6.9$ & $72.9\pm6.3$ & $-7.27\pm0.19$ & $0.104\pm0.003$\\
Full
& $\mathbf{94.1\pm1.5}$ & $\mathbf{82.5\pm3.2}$ & $\mathbf{-7.85\pm0.01}$ & $\mathbf{0.094\pm0.002}$\\
\hline
\end{tabular}}
\end{table}

\begin{figure}[t]
\centering
\includegraphics[width=0.85\linewidth]{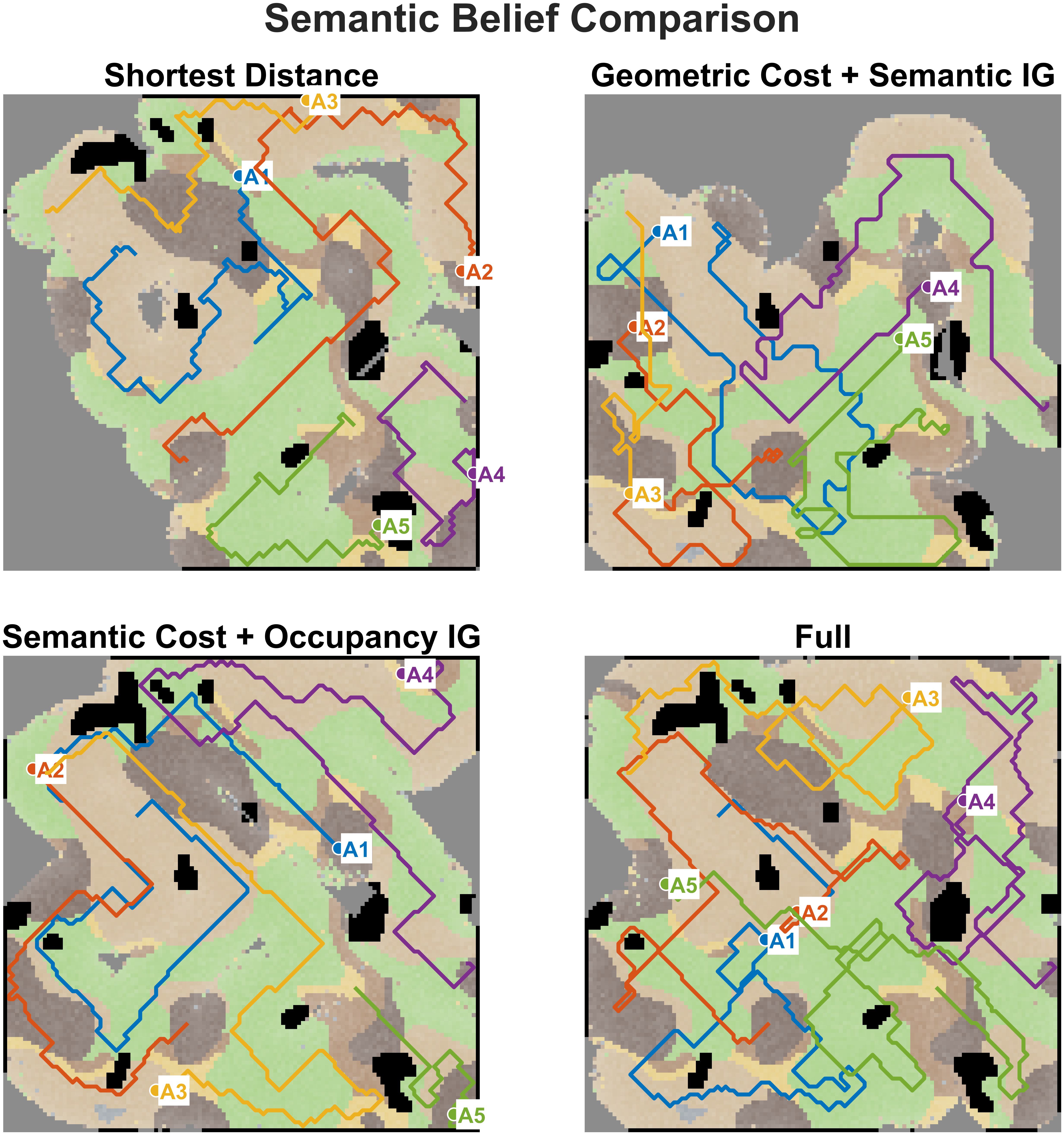}
\caption{Final fused semantic maps for a representative paired run after 300~s. Colored lines show robot trajectories, black cells are obstacles, and gray cells are unexplored.}
\label{fig:matlab_sem_beliefs}
\end{figure}

As shown in Table~\ref{tab:ablation}, Full improves coverage and mIoU over Shortest Distance by $14.3$ and $14.2$ percentage points, respectively, while achieving the lowest entropy and route cost. These improvements are statistically significant ($p<0.01$). Figure~\ref{fig:matlab_sem_beliefs} shows a representative paired run. Both partial ablations improve the mean map-quality metrics, although their coverage and mIoU improvements over Shortest Distance are not statistically significant. The lower route costs of the terrain-aware modes indicate that semantic traversal costs direct robots toward easier terrain.

\subsection{Mobility-Budget Sensitivity}

We evaluate the full framework over budgets $B\in\{0.75,1.0,\ldots,2.75\}$ using 10 paired runs per setting. As shown in Table~\ref{tab:budget}, performance is non-monotonic: $B=1.25$ achieves the highest mean coverage ($95.0\%$) and mIoU ($85.4\%$). Increasing the budget expands the feasible trajectory set and reduces the progress-fallback rate from $100\%$ at $B=0.75$ to $3.3\%$ at $B=2.75$. However, smaller budgets frequently require budget-feasible progress waypoints, while larger budgets produce longer trajectories whose predicted information gain may not be fully realized before receding-horizon replanning. These results demonstrate that the budget actively shapes trajectory selection and that an intermediate value best balances immediate reachability with achievable information gain.

\begin{table}[t]
\centering
\caption{Mobility-budget sensitivity over 10 runs (mean $\pm$ 95\% CI).}
\label{tab:budget}
\begin{tabular}{cccc}
\hline
$B$ & Coverage (\%) $\uparrow$ & mIoU (\%) $\uparrow$ & Fallback (\%) $\downarrow$\\
\hline
$0.75$ & $91.3 \pm 3.6$ & $80.8 \pm 4.3$ & $100.0$\\
$1.00$ & $94.1 \pm 1.5$ & $82.5 \pm 3.2$ & $100.0$\\
$\mathbf{1.25}$ & $\mathbf{95.0 \pm 2.1}$ & $\mathbf{85.4 \pm 2.8}$ & $76.0$\\
$1.75$ & $93.7 \pm 2.0$ & $82.1 \pm 3.1$ & $17.5$\\
$2.25$ & $89.8 \pm 2.3$ & $78.3 \pm 1.9$ & $7.2$\\
$2.75$ & $81.5 \pm 3.1$ & $68.8 \pm 4.0$ & $3.3$\\
\hline
\end{tabular}
\end{table}

\begin{figure}[t]
\centering
\includegraphics[width=0.85\linewidth]{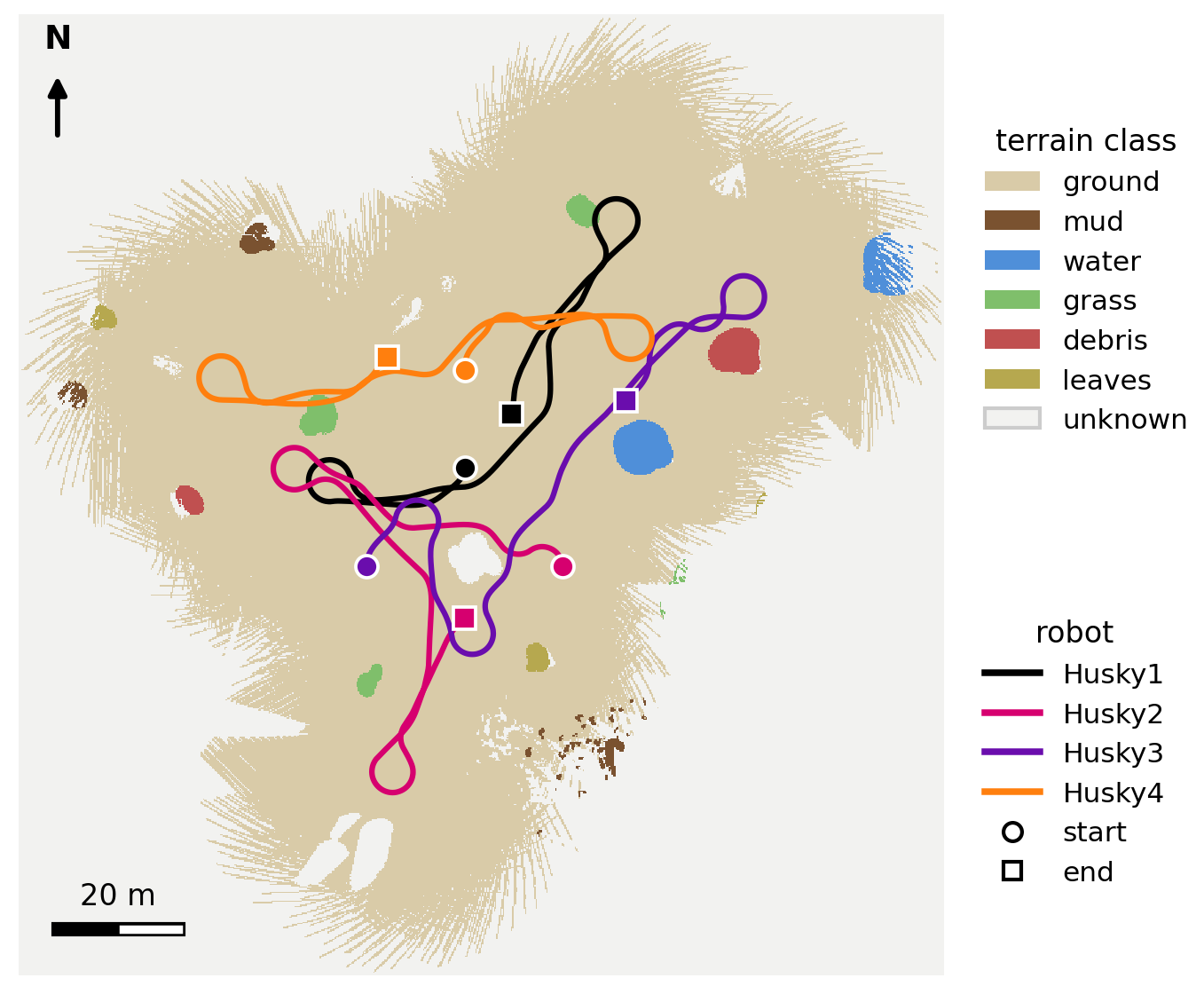}
\caption{Fused semantic ground map and robot trajectories after a 600~s run with four ground robots. Circles mark start poses, squares mark final poses.}
\label{fig:hercules}
\end{figure}

\subsection{Full-System Evaluation}

We evaluate the complete framework with four Husky ground robots in HERCULES \cite{hercules}, an open-source, photorealistic outdoor simulator. Each robot builds a 3-D semantic voxel map from depth, segmentation, and pose collected in lockstep under a single simulation timestamp, derives a 2-D Dirichlet semantic grid by per-column clearance analysis, and runs the framework of Section~\ref{sec:method}. A communication broker enforces the configured topology on every exchanged source-attributed record, bid, and commitment. Fig.~\ref{fig:hercules} shows a representative run.

Terrain classes carry traversal costs $S=[3,1,8,6,25,4.5,7,4]$ for ground, road, rough, mud, water, grass, debris, and leaves, with matching speed caps enforced by a terrain governor so that difficult terrain also costs mission time. The heading penalty uses $\lambda=0.3$ with turning angles normalized by $\pi$, and the Dirichlet prior is $\alpha_0=1$ over $K=8$ classes. Frontier cells are clustered into 8-connected components; clusters smaller than five cells are discarded, and each remaining cluster is represented by an adjacent traversable cell. The per-round budget $B=180$ in native planner cost units is calibrated so that roughly 40\% of frontier candidates are infeasible. A continuous frozen-robot detector rejects runs in which any robot becomes immobile.

\textbf{Baselines.} \emph{Independent} selects greedily without coordination. \emph{Voronoi} assigns frontiers to their nearest robots. \emph{Auction} sequentially assigns the highest-bidding robot-candidate pair. \emph{NBV-3DMR} follows \cite{freda3dmr}, scoring information gain exponentially discounted by path length and resolving conflicts in favor of the lower-cost path. \emph{ROAM-overlap} follows \cite{asgharivaskasi2025riemannian}, penalizing neighboring sensor-footprint overlap until selections stabilize. We port only the coordination rules because the complete pipelines would confound coordination with mapping and planning. All methods use the shared Dirichlet grid, with semantic path costs from \eqref{eq:cost} in the coordination ablation and geometric path length in the native-form comparison.

\textbf{Coordination ablation.} When all six modes share the same semantic map, traversal costs, budget-feasible candidates, and objective, their coverage, coverage per meter, mIoU, entropy, and budget-infeasible fractions are statistically indistinguishable over three 600-s runs: all pairwise 95\% confidence intervals overlap. Coordination provides the only separable effect. All five coordinated modes reduce same-target selections to approximately zero per round, compared with $0.196$ for Independent (95\% confidence interval $0.093$--$0.300$). This indicates that the shared framework dominates aggregate mapping performance, motivating comparison of the coordination rules in their native published forms.

\begin{table}[t]
\caption{Native-form comparison in HERCULES over five 600-s runs per method (mean $\pm$ 95\% confidence interval).}
\label{tab:native}
\centering
\scriptsize
\setlength{\tabcolsep}{4pt}
\renewcommand{\arraystretch}{1.15}
\begin{tabular}{lccc}
\hline
\textbf{Method} &
\shortstack{\textbf{Coverage}\\($10^3$ cells)} &
\textbf{mIoU} &
\shortstack{\textbf{Budget}\\\textbf{violation}} \\
\hline
Semantic-RAG (ours)
& $149.3 \pm 15.5$
& $0.426 \pm 0.037$
& $\mathbf{0}$ \\
Voronoi
& $142.6 \pm 12.8$
& $0.396 \pm 0.046$
& $0.854 \pm 0.054$ \\
NBV-3DMR
& $131.7 \pm 16.9$
& $0.402 \pm 0.020$
& $0.974 \pm 0.014$ \\
Auction
& $127.1 \pm 14.6$
& $0.404 \pm 0.138$
& $0.969 \pm 0.006$ \\
ROAM-overlap
& $133.8 \pm 21.2$
& $0.403 \pm 0.162$
& $0.951 \pm 0.034$ \\
Independent
& $130.3 \pm 12.2$
& $0.383 \pm 0.102$
& $0.977 \pm 0.013$ \\
\hline
\end{tabular}
\end{table}

\textbf{Native-form comparison.} Each baseline is rerun in its published form: no semantic traversal cost model and, for all methods other than ours, occupancy-only information gain and no per-round mobility budget. All methods share the same mapper, planner, executor, and sensing system. Coverage and mIoU are measured at energy exhaustion, while the baseline budget-violation rate is the fraction of committed trajectories exceeding $B$. As shown in Table~\ref{tab:native}, the proposed method satisfies the per-trajectory budget by construction, whereas the baselines exceed it on 85\%--98\% of committed trajectories. It also attains the highest mean coverage and mIoU, although the confidence intervals overlap, so these are consistent trends rather than statistically separated differences.

\section{Conclusion}

We presented a distributed semantic multi-robot exploration framework combining Dirichlet mapping, budget-constrained trajectory selection, and RAG coordination, with exact per-robot selection, a $1/2$ joint per-round guarantee, and communication-dependent bounds. Gridworld ablations show that semantic information gain and semantic traversal costs contribute complementarily. In full-system simulation, holding the framework fixed shows that coordination of any kind eliminates redundant target selection and that the framework rather than the selection rule dominates task performance at this sample size, while running the baselines in their native published forms under a shared mission energy budget leaves the proposed method with the highest mean coverage and semantic accuracy and as the only one that satisfies the per-trajectory mobility budget.

Limitations are the small sample sizes, the absence of a sensitivity analysis over the terrain cost vector, heading penalty, and Dirichlet prior, a single sensor accuracy setting, and validation in simulation only. Future work adds hardware deployment, evaluation across communication topologies, and the full metric suite, per-class accuracy, and energy-versus-coverage analysis reported in the journal version.

\bibliographystyle{IEEEtran}
\bibliography{IEEEabrv,ref}

\end{document}